\documentclass{sig-alternate-05-2015}
\pdfoutput=1

\usepackage{supertabular}
\usepackage{booktabs}
\usepackage{needspace}

\providecommand{\tightlist}{%
  \setlength{\itemsep}{0pt}\setlength{\parskip}{0pt}}

\begin{document}

\title{cltorch: a Hardware-Agnostic Backend for the Torch Deep Neural Network Library, Based on OpenCL}

\numberofauthors{1} 
\author{
\alignauthor
Hugh Perkins\\
       \affaddr{ASAPP}\\
       \email{hp@asapp.com}
}

\maketitle
\begin{abstract}

This paper presents \texttt{cltorch}, a hardware-agnostic backend for
the Torch neural network framework. \texttt{cltorch} enables training of
deep neural networks on GPUs from diverse hardware vendors, including
AMD, NVIDIA, and Intel.

\texttt{cltorch} contains sufficient implementation to run models such
as AlexNet, VGG, Overfeat, and GoogleNet. It is written using the OpenCL
language, a portable compute language, governed by the Khronos Group.
\texttt{cltorch} is the top-ranked hardware-agnostic machine learning
framework on Chintala's convnet-benchmarks page.

This paper presents the technical challenges encountered whilst creating
the \texttt{cltorch} backend for Torch, and looks in detail at the
challenges related to obtaining a fast hardware-agnostic implementation.

The convolutional layers are identified as the key area of focus for
accelerating hardware-agnostic frameworks. Possible approaches to
accelerating the convolutional implementation are identified including:

\begin{itemize}
\tightlist
\item
  implementation of the convolutions using the \texttt{implicitgemm} or
  \texttt{winograd} algorithm
\item
  using a GEMM implementation adapted to the geometries associated with
  the convolutional algorithm
\item
  using a pluggable hardware-specific convolutional implementation
\end{itemize}

\end{abstract}

\printccsdesc


\section{Introduction}\label{introduction}

This work presents an OpenCL backend for the Torch network library.
Torch\cite{torch} is a deep learning library. Other commonly used deep
learning libraries include: caffe\cite{caffe}, MXNet\cite{mxnet}, and
Tensorflow\cite{tensorflow}.

Deep learning became popular with the success of the
AlexNet\cite{alexnet} deep learning model on the Imagenet\cite{imagenet}
task. Prior to AlexNet, image recognition used hand-engineered features
such as SIFT\cite{sift} and SURF\cite{surf}, in order to get state of
the art performance. AlexNet used deep learning to extract features
automatically, via the expressive power of a hierarchy of layers. A key
ingredient for this to work was the massive amount of data available in
the ImageNet task, compared to earlier tasks, such as MNIST\cite{mnist}
and CIFAR\cite{cifar}. This allowed training complex models without
overfitting. ImageNet gave sufficient training data for the gradients to
propagate across all layers, despite the vanishing of the gradient
signal as it traverses from the output back through the multiple layers.
Given the amount of training data involved, and the length of time to
train each batch in a deep network, another key ingredient for the
success of AlexNet was: the computational capacity available in GPUs.
GPUs provided sufficient processing power to be able to run multiple
epochs on imagenet, using a deep network, in a reasonable timeframe, ie
days or weeks; rather than months or years.

Recent continued developments of SOTA models on ImageNet often improve
on AlexNet by adding more layers, making the network deeper. It is
probably fair to say that the more processing power we can bring to bear
onto neural network models, the easier it will be to create powerful
models, to train them in a reasonable timeframe, and to train them on
ever larger datasets. LSUN dataset\cite{lsun} for example has 20 million
images, a 20-fold increase from AlexNet.

In order to obtain continued development of the computational capacity
of neural network hardware, it could plausibly be useful to encourage
competition between hardware vendors to as great an extent possible. For
training neural networks, GPUs are typically used, but there is no
particular reason why other hardware, such as Field Programmable Gate
Arrays (``FPGAs''), or dedicated Massively Parallel Processors (MPPs),
could not be used instead. Even within the domain of GPUs, it is
probably fair to say that the vast majority of neural network training
is not done on multiple hardware vendor's hardware, but using NVIDIA
CUDA hardware. One reason for this is that NVIDIA was the first vendor
to make GPUs accessible for running numerical algorithms in a
straightforward fashion, without needing clever hardware hacks. NVIDIA
introduced the CUDA toolkit, and they released TESLA GPUs, which were
dedicated hardware for running embarrassingly parallel numerical
algorithms. More recently, other GPU manufacturers, such as Intel and
AMD, have made their own GPUs available for numerical compute
applications. However, looking at the machine learning domain, existing
neural network libraries were written to use the NVIDIA CUDA toolkit,
and it was not therefore possible to run them on non-CUDA hardware.
Concretely it is not possible to run Caffe, Tensorflow, Torch, or MXNET
on non-CUDA hardware.

In 2015, this situation started to change. Caffe now can be used on
OpenCL hardware, thanks to the efforts of Gu et al \cite{openclcaffe},
Tschopp\cite{greentea}, and Engel\cite{caffeclengel}.
DeepCL\cite{deepcl} provides a dedicated deep learning library for
OpenCL. \texttt{cltorch}, described in the present paper, provides a
hardware-agnostic backend for the `Torch' neural network library, using
OpenCL. These projects facilitate vendor competition, and therefore
contribute to the development of state of the art parallel compute
hardware.

\texttt{cltorch} is built using OpenCL. OpenCL is a standard governed by
the Khronos Group. It aims to make it possible to run a single numerical
algorithmic implementation across multiple hardwares, without needing
any changes to the implementation. Hardware vendors who are members of
the OpenCL consortium include: AMD, Qualcomm, NVIDIA, Intel, and Xilinx.

OpenCL is not the only possible cross-platform approach possible. A
recent alternative is AMD's HCC, which we will touch on briefly, at the
end of this paper.

\section{Key differences between CUDA and
OpenCL}\label{key-differences-between-cuda-and-opencl}

CUDA and OpenCL are not so different, both conceptually, and in the
details. However there are some key differences, that make porting
challenging.

\subsection{language word differences}\label{language-word-differences}

The names of certain functions are different. For the example, to
determine the id of a thread within a block of threads, on CUDA this
uses properties, such as \texttt{threadIdx.x}. In OpenCL this uses
functions such as \texttt{get\_local\_id(0)}. The meanings of such
functions are typically the same, or similar, and these differences can
be handled in a straightforward fashion, using search and replace.

\subsection{C++ templates}\label{c-templates}

A more fundamental difference is the use of C++, and specifically C++
templates and functors, in CUDA. Torch CUDA framework makes extensive
use of these. OpenCL 1.2 is based on C99, and does not have templating.
Gu et al solved this problem by using a recent version of OpenCL, which
does support C++ templates. Unfortunately, this limits the hardware that
such a library can run on. For example NVIDIA hardware, and some mobile
hardware, only supports OpenCL 1.1 and OpenCL 1.2. It can be argued that
there is no reason to support NVIDIA hardware, because on NVIDIA
hardware one can use CUDA instead. However, it seems an interesting goal
to be able to run a neural network library across all commonly available
GPUs, including those for which an alternative implementation already
exists. This would enable a single codebase to be used across all
devices, reducing redundant effort.

The C++ templates in Torch CUDA implementation cannot be trivially
replaced by hand-written C99 code. For example, they are used to
pre-bake geometry-specific parameters, such as loop sizes for loop
unrolling. It is not tempting to write unrolled loops by hand for
multiple possible geometries. The clBLAS\cite{clblas} library handles
this issue by using Python scripts to directly write out the generated
OpenCL kernels and related C code, at compile time. This is an effective
solution. It is fast at runtime, because the generated code is entirely
generated at build time. However, it means that all possible geometries
need to be baked in, at compile time, leading to a combinatorial
explosion of possible parameter values.

Therefore, in \texttt{cltorch}, the kernels are generated at runtime,
adapted to the exact geometries required. Runtime code generation needs
a scripting language to be embedded into the network library. Lua was
selected. It is lightweight (about 46KB), and easy to embed. A
templating language similar to Jinja2 was used, based on a templater by
Nachimwald\cite{nachiwald}. This approach was sufficient to express the
C++ CUDA templates used by CUDA torch.

\subsection{Use of Thrust library}\label{use-of-thrust-library}

The Torch CUDA library makes extensive use of the Thrust\cite{thrust}
library, to handle for example sorting and reduction operations. Thrust
is CUDA-specific, and therefore these methods need careful
consideration. For now, this was worked around by using alternative
approaches. In the case of reduction, Torch CUDA backend has internal
implementations, and this was ported, and used in place of some of the
calls to Thrust. Merry \cite{merry2015performance} provides performance
benchmarks of several libraries that might be useful here.

\subsection{cl\_mem object offsets}\label{clux5fmem-object-offsets}

cl\_mem objects cannot themselves carry an implicit offset within the
allocated buffer, unlike their CUDA \texttt{float\ *} counterparts. In
Caffe library OpenCL backends, this was quite a large engineering
challenge, since the \texttt{float\ *}s are passed around inside the
library, and offsets are added to these in very many places. In Torch
OpenCL backend this was not an issue, since the Torch Tensor structure
already incorporates a \texttt{storageOffset} value.

\subsection{Runtime compilation}\label{runtime-compilation}

In order to facilitate being able to run across many hardware platforms,
OpenCL kernels are typically compiled on the fly, at runtime. CUDA
kernels are compiled at build time, with the rest of the C and C++ code.
Compiling the GPU kernels at build time might provide a very slight
boost at runtime, but in practice the cost of compiling the OpenCL
kernels at runtime was found to be negligible, for deep learning
libraries, compared to the time spent on network training. Compiling at
runtime has the advantage that one can bake in the exact geometries
which are required. By comparison, many of the Torch CUDA functions,
such as the `Apply' method for example, bake in several hundred possible
geometries, most of which will never be used. Runtime compilation makes
it possible to provide some useful functionalities to the framework
user, such as being able to provide custom functions, which will be
compiled on the fly into GPU kernels. These will run on the GPU, at full
speed. Zagoruyko has implemented similar functionality for CUDA Torch as
an additional module \cite{cunnrtc}, which runs the CUDA compiler at
runtime.

\subsection{OpenCL Kernel Debuggers}\label{opencl-kernel-debuggers}

When using CUDA, there are extensive debugging tools available, to
facilitate finding root cause on kernel crashes, or numerical accuracy
issues. When using OpenCL on CUDA devices, no such debugging tools are
available. Some vendors provide debugging tools for their specific
hardware. For example AMD has created CodeXL\cite{codexl}. However, no
cross-platform tool is available. An alternative approach is to use a
simulator. Oclgrind \cite{oclgrind} provides OpenCL kernel debugging,
using a simulated OpenCL device, loadable via ICD.

\subsection{Hardware geometry
differences}\label{hardware-geometry-differences}

Each specific hardware has its own set of geometry specifications,
including, but not limited to:

\begin{itemize}
\tightlist
\item
  maximum workgroup size
\item
  warp size
\item
  local memory size
\item
  number of registers
\end{itemize}

A kernel that runs prefectly on NVIDIA hardware might fail to run
altogether on eg AMD hardware. For example, the kernel might be written
on the assumption that at least 512 workgroup threads are available,
whereas AMD hardware typically has a maximum of 256 workgroup threads.
Many kernels won't run across multiple hardware platforms until they
have been tested on each platform, and platform-specific issues
addressed. It is not in general safe to assume that code that works on
one hardware platform will run on all the others.

Moreover, the low-level memory management involved in writing OpenCL
code contrasts with the use of caching on CPUs, where the caching is
handled automatically by the processor, at runtime. It arguably might
not be appropriate for GPUs to handle caching dynamically at runtime,
because one of the distinguishing features of GPU cores is their
simplicity. However, it does seem plausible that a language that is
slightly higher level than OpenCL could perhaps leave the management of
the cache and the memory to an effective optimizing compiler to handle?
This would free the developer to focus on writing down the algorithms in
programmatic form, rather than considering different possible
permutations and combinations of memory and loop structures. PENCIL
\cite{pencil} might be one approach here.

\subsection{Compiler differences}\label{compiler-differences}

Each hardware vendor's compiler emits slightly different warnings and
errors. Something that builds cleanly on one vendor's hardware might
fail to build altogether on another's. In addition, certain syntax,
valid C99, entirely crashed certain compilers, a ``segfault''. The
result is that, in addition to hardware geometry differences, each
kernel needs to be built using each supported vendor's OpenCL compiler,
in order to resolve vendor-specific build issues. It is not safe to
assume that if it builds on one vendor's OpenCL compiler, and that the
kernel contains only valid C99, that other vendor's OpenCL compilers
will build the kernel successfully.

Therefore, if one wishes to support hardware from three different
vendors, one needs to have access to representative hardware from each
vendor.

\subsection{OpenCL version}\label{opencl-version}

There is a compromise between targeting more recent versions of OpenCL,
providing additional functionality and performance, or targeting older
versions, allowing the use of more possible target platforms.
\texttt{cltorch} targets OpenCL 1.2, which is supported by many current
GPUs from Intel, NVIDIA, and AMD. Nevertheless, some mobile devices in
common use continue to use OpenCL 1.1, and would therefore need special
treatment in order to work with \texttt{cltorch}.

\subsection{Kernel caching}\label{kernel-caching}

NVIDIA GPUs cache the kernels automatically, even when building from
templates, including where the kernel OpenCL code is supplied as a
string, at runtime. For training a neural network model, the compilation
time is negligible. However, for development purposes, being able to
start the kernels quickly is convenient. Thus, there could be an
opportunity for other vendors to provide similar kernel caching
functionality.

\section{Comparison with other
frameworks}\label{comparison-with-other-frameworks}

In this section, we will compare OpenCL Torch with other deep learning
frameworks and backends, and specifically with:

\begin{itemize}
\tightlist
\item
  CUDA Torch
\item
  DeepCL
\item
  OpenCL Caffe
\end{itemize}

\subsection{CUDA Torch}\label{cuda-torch}

CUDA Torch provides more functionality than the OpenCL Torch
implementation currently. For example, the \texttt{sort} function is not
currently implemented in OpenCL Torch.

\needspace{36pt}Given constant hardware, the CUDA implementation runs
faster than the OpenCL implementation. There are two principle reasons:

\begin{itemize}
\tightlist
\item
  Semantically identical OpenCL kernels often run slower than their CUDA
  counterparts, even on the same hardware
\item
  there is currently no equivalent of the CUDA CUDNN\cite{cudnn} library
  available for OpenCL
\end{itemize}

\begin{figure}[htbp]
\centering
\includegraphics[width=0.48000\textwidth]{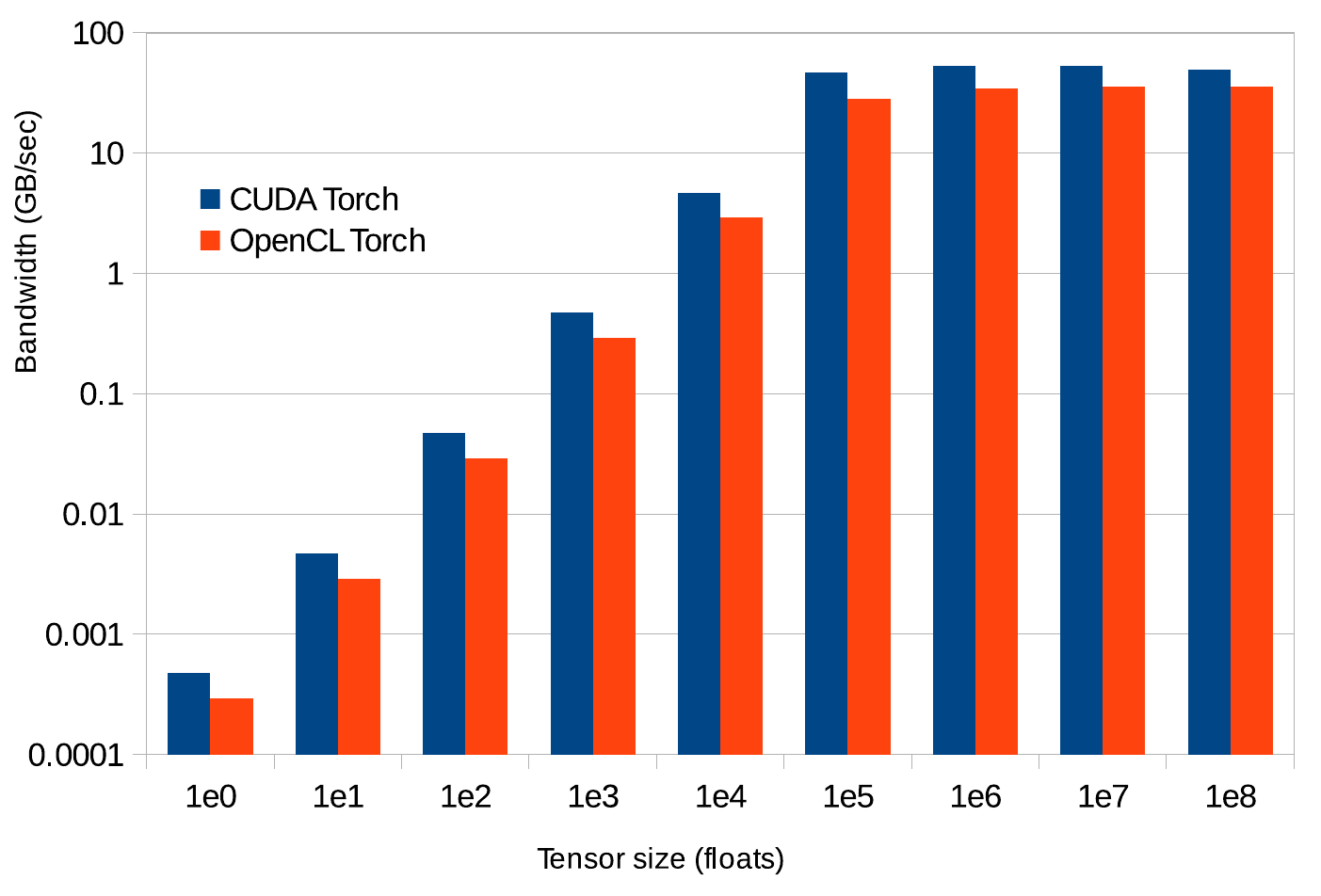}
\caption{Per-element timings as function of tensor
size\label{perelement1}}
\end{figure}

With regards to the first reason, Figure \ref{perelement1} shows a
comparison of carrying out per-element operations on a tensor in CUDA
Torch vs OpenCL Torch. It shows the bandwidth for per-element
operations, as a function of the number of floats per kernel launch, for
CUDA and for OpenCL. These kernels are relatively simple, and the code
is more or less the same, to within a search and replace eg of
\texttt{threadIdx.x} with \texttt{get\_local\_id(0)}. This experiment
was performed on an NVIDIA K520 GPU, using CUDA Torch as of 21 May 2016,
and OpenCL Torch as of 21 May 2016. The scale is log-log. We can see
that the shape of the graph is pretty similar for both OpenCL and CUDA
kernel launches. For both backends, the bandwidth is approximately
constant down to tensors of \(1e5\) floats, then reduces with tensor
size, as the overhead of kernel launch and setup becomes dominant. The
shape of both graphs, CUDA vs OpenCL, on a log-log scale, looks pretty
similar.

\begin{figure}[htbp]
\centering
\includegraphics[width=0.50000\textwidth]{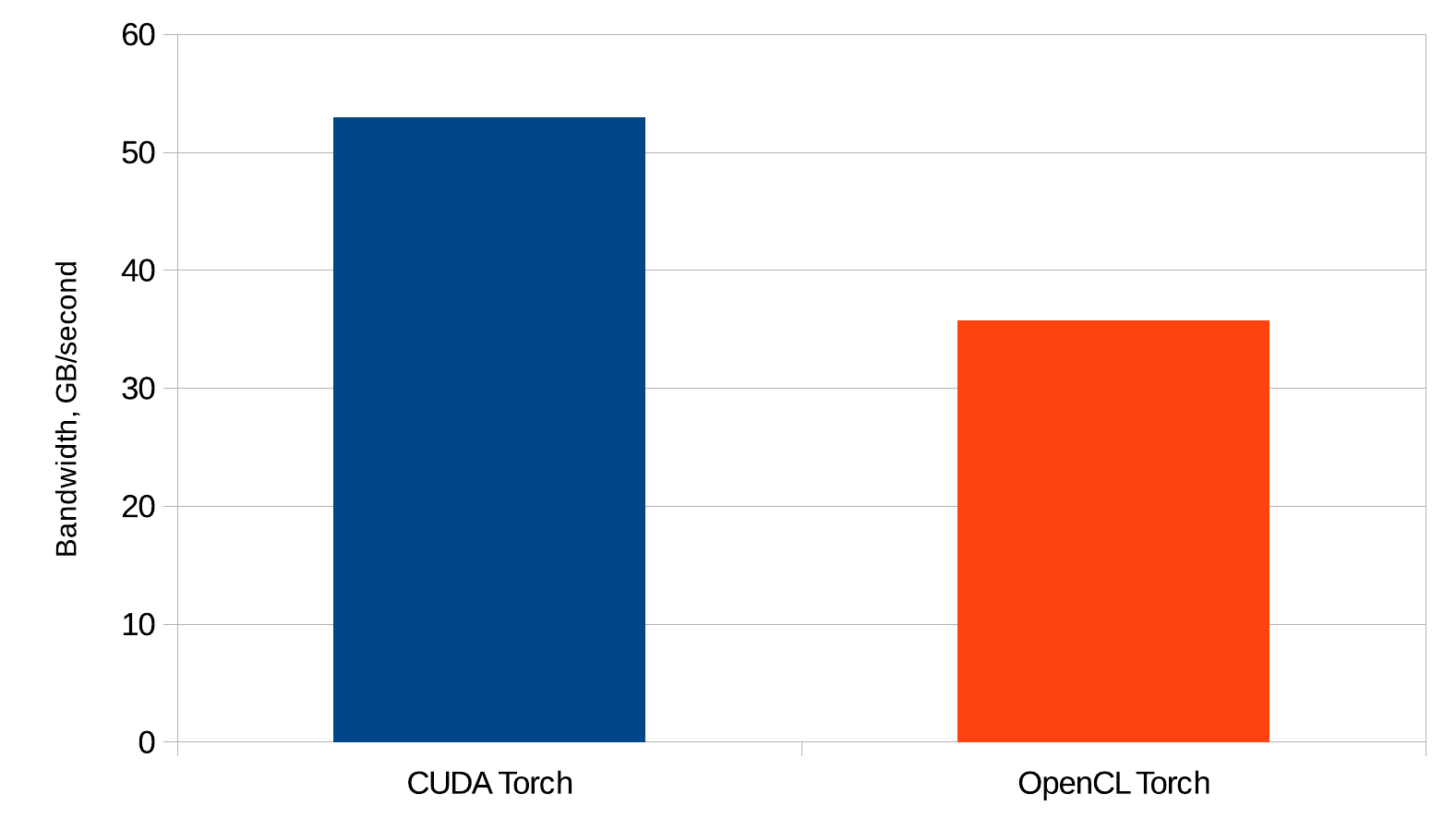}
\caption{Per-element timings for tensors of 1e7
floats\label{perelement2}}
\end{figure}

However on a linear scale, there is a noticeable difference between the
performance of this kernel using CUDA and using OpenCL. Figure
\ref{perelement2} shows the bandwidths on a linear scale for tensors of
\(1e7\) floats.\\
We can see that the OpenCL backend is about 33\% slower than the CUDA
backend, for this geometry, even though the hardware is identical, and
the kernel code is approximately identical, to within a search and
replace for eg \texttt{threadIdx.x} vs \texttt{get\_local\_id(0)}. An
opportunity exists to look at why this is the case, since both CUDA and
OpenCL compile via the same PTX IL.

\subsection{DeepCL}\label{deepcl}

DeepCL\cite{deepcl} provides a deep learning framework dedicated to
OpenCL, that runs on Windows, as well as on linux and Mac. It provides a
commandline tool, and a python interface.

DeepCL does not provide as many network layer types, or as complete an
implementation of each layer type, compared with OpenCL Torch, or OpenCL
Caffe.

\subsection{OpenCL Caffe}\label{opencl-caffe}

There are several ports of Caffe\cite{caffe} to OpenCL: Gu et
al\cite{openclcaffe}, Engel\cite{caffeclengel}, and
Tschopp\cite{greentea}

Caffe is quite challenging to port to OpenCL, because it is quite
tightly coupled with the underlying CUDA implementation, and it is
challenging to factorize this so that an OpenCL implementation will not
adversely affect subsequent Caffe development. There is no easy solution
for this task, so merging the OpenCL ports to Caffe core is a
painstaking time-consuming task, needing careful negotiation between the
various parties involved.

In terms of performance, the port by Gu et al contains a batched
\texttt{im2col} implementation, discussed below. Tschopp's Greentea
initially used \texttt{im2col}. Recently, Tschopp has started to
implement \texttt{implicitgemm}. These algorithms are discussed below.

Having discussed some alternative implementations in terms of features
and functionality, let's look now at performance.

\subsection{Performance}\label{performance}

\begin{figure}[htbp]
\centering
\includegraphics[width=0.48000\textwidth]{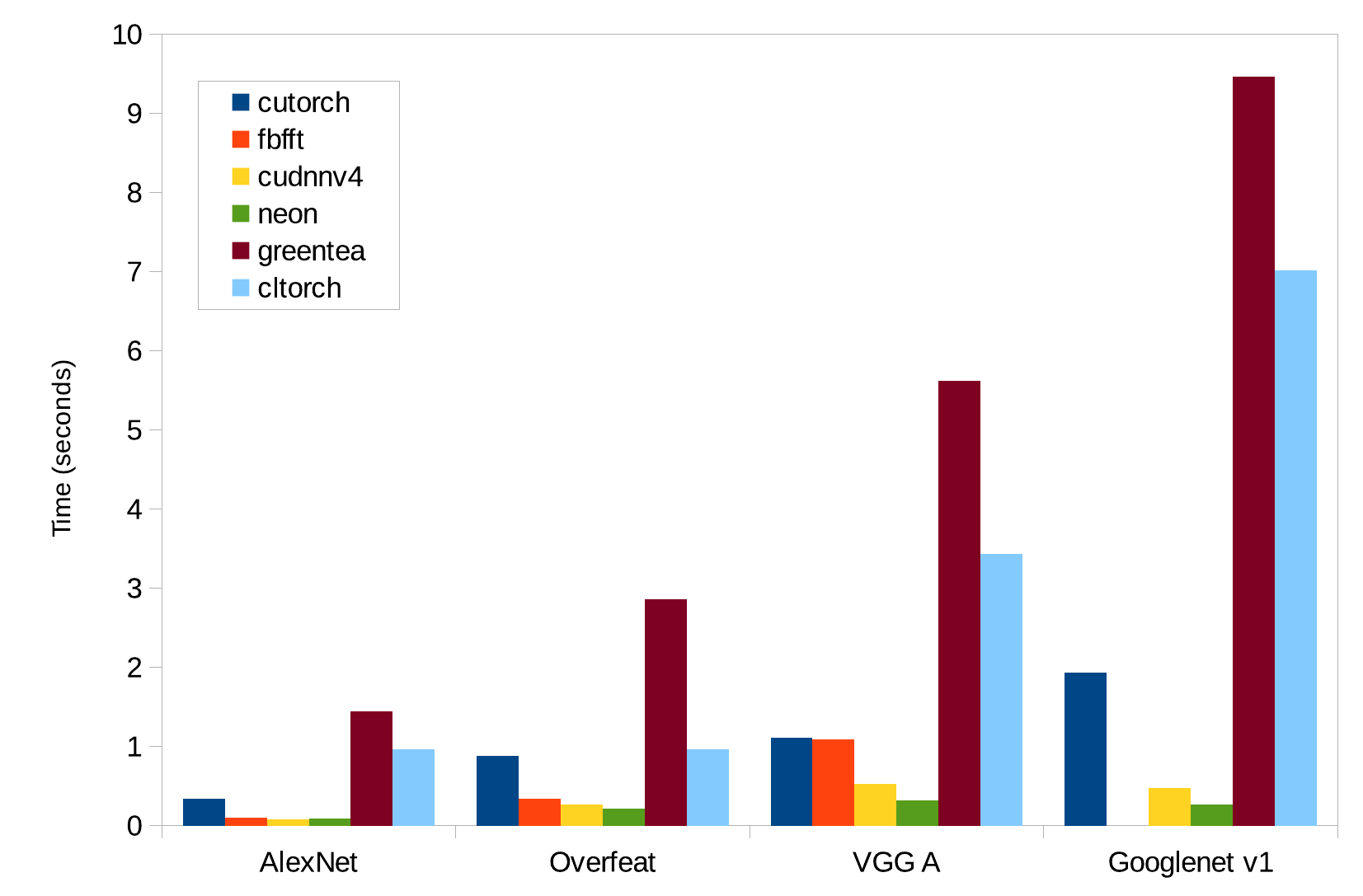}
\caption{Convnet-benchmarks timings\label{convnetbenchmarks}}
\end{figure}

Figure \ref{convnetbenchmarks} shows the timings for running a single
forward-backward iteration through the convolutional layers of six
popular convnet models. These timings are from Chintala's
convnet-benchmarks\cite{convnetbenchmarks}. The timings shown are the
forward-backward training times for a single batch. The implementations
are described in Table \ref{tab:convnetimplementations}.

\begin{table}
\begin{center}
\begin{tabular}[h]{llll}
\toprule
Identifier & Framework & Backend & Algorithm \\ \hline
\texttt{cutorch} & Torch & CUDA & \texttt{im2col} \\
\texttt{fbfft} & Torch & CUDA & \texttt{fft} \\
\texttt{cudnnv4} & Torch & CUDA & CUDNNv4 \\
\texttt{neon} & Nervana Neon & CUDA & \texttt{winograd} \\
\texttt{greentea} & Caffe & OpenCL & \texttt{im2col} \\
\texttt{cltorch} & Torch & OpenCL & \texttt{im2col} \\
\bottomrule
\end{tabular}
\end{center}
\caption{Convolutional Implementations}
\label{tab:convnetimplementations}
\end{table}

Thus, we compare between torch implementations, between OpenCL Torch and
OpenCL Caffe, and with a state of the art convolutional implementation
for CUDA cards \texttt{neon}. DeepCL is included in the per-layer
timings on the convnet-benchmarks page, but is not included in the full
network timings shown here, because it lacks certain key implementation
details for these models. Looking at Figure \ref{convnetbenchmarks}, we
can see that the OpenCL implementations are significantly slower than
the CUDA implementations. \texttt{cltorch} is marginally faster than
Tschopp's Greentea \texttt{im2col} implementation. \texttt{cltorch} is
around 3-4 times slower than the Torch \texttt{cutorch} implementation.
\texttt{neon} is about 20 times faster than \texttt{cltorch}, on the
\texttt{Googlenet\ v1} model. Gu et al's OpenCL Caffe is not present on
the benchmarks page, because it uses AMD-specific OpenCL extensions, and
therefore cannot run on the NVIDIA Titan X used here. It could be
interesting to obtain benchmarks on hardware from other vendors, for
example on AMD GPUs.

Let's look at performance in more detail, and look at what options we
have to improve the performance of convolutions in hardware-agnostic
neural network frameworks.

\section{Performance Analysis}\label{performance-analysis}

In this section, we will analyse options for improving the performance
of hardware-agnostic neural net frameworks. We will see that the
convolutional layers dominate the training time. We will compare common
algorithms for computing the convolution, discuss GEMM,
hardware-specific implementations, and HCC.

\subsection{Importance of convolution}\label{importance-of-convolution}

\begin{figure}[htbp]
\centering
\includegraphics[width=0.48000\textwidth]{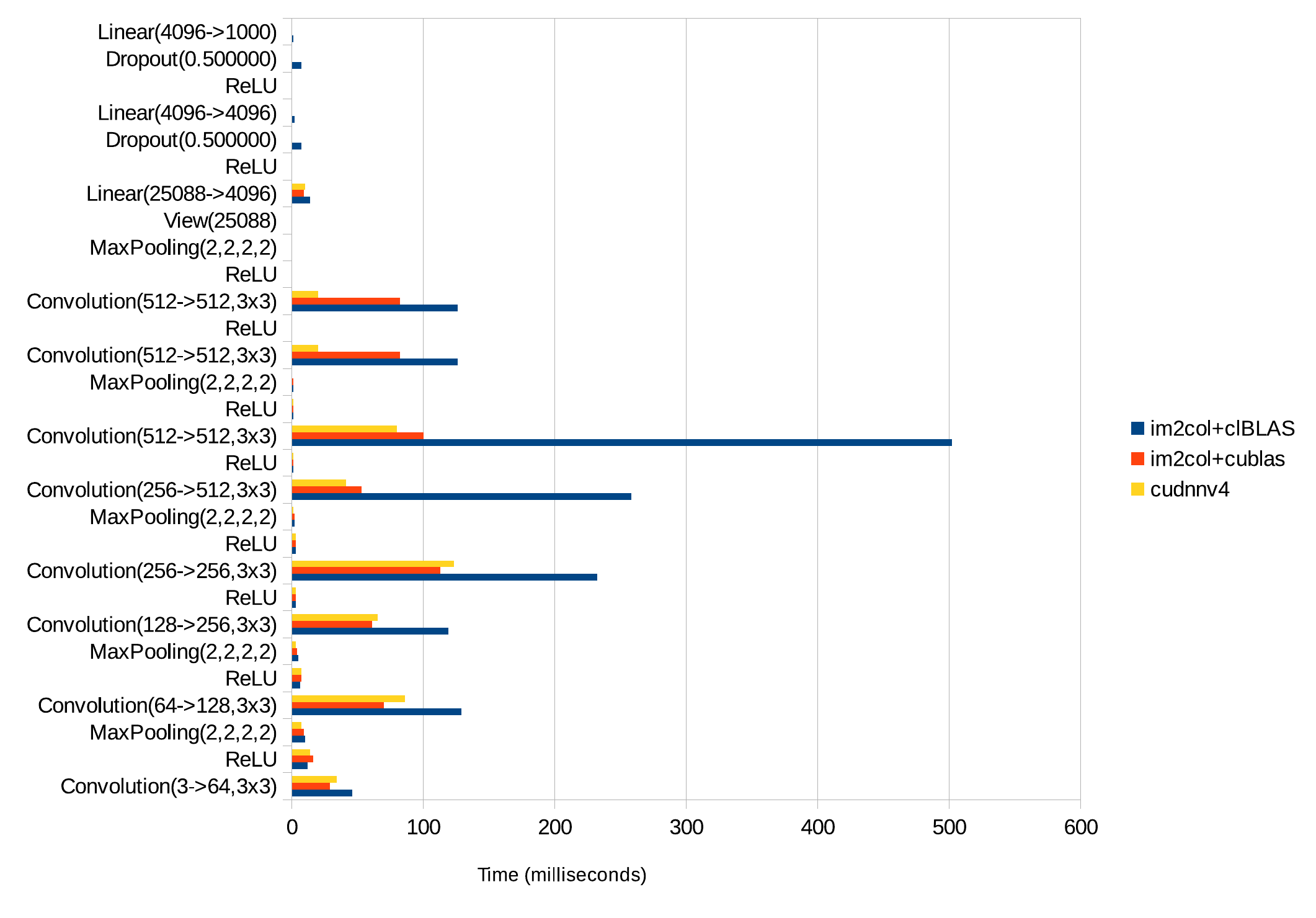}
\caption{VGG A per-layer timings\label{vggalayertimings}}
\end{figure}

The training time for a convolutional network is dominated by the
convolutional layers. Figure \ref{vggalayertimings} shows the per-layer
forward time for VGG model `A', on an NVIDIA Titan X GPU, for
\texttt{im2col+clBLAS}, \texttt{im2col+cublas} and \texttt{cudnnv4}
convolutional implementations.

\begin{figure}[htbp]
\centering
\includegraphics[width=0.48000\textwidth]{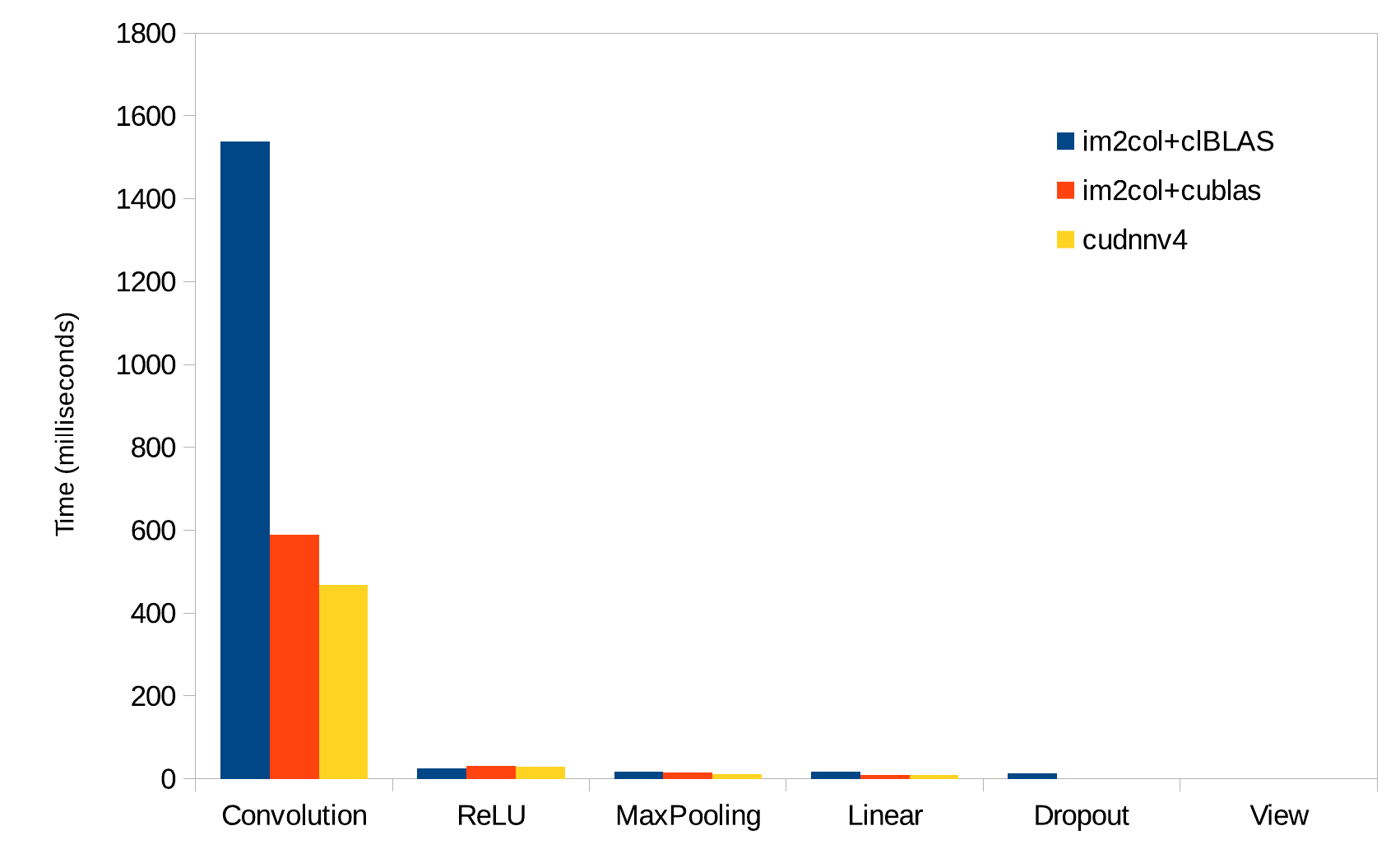}
\caption{VGG timings grouped by layer
type\label{vggalayertimingssummary}}
\end{figure}

The vast majority of the time is going into the convolutional layers,
rather than into pooling, or RELU. Figure \ref{vggalayertimingssummary}
sums the timings by layer type, and confirms this analysis. It is clear
that the convolutional layer is where we should be focusing our
attention.

These graphs shows timings for three Torch GPU backends, described in
Table \ref{tabletorchbackends}. Looking again at Figure
\ref{vggalayertimingssummary}, we can see that the convolutional
performance of \texttt{cltorch} is about three times slower than for the
CUDA backends for Torch, and therefore there are opportunities to
improve in this area.

\begin{table}
\begin{center}
\begin{tabular}[h]{ll}
\toprule
Implementation & Torch GPU backend \\ \hline
\texttt{im2col+cublas} & \texttt{cutorch} (CUDA Torch) \\
\texttt{cudnnv4} & \texttt{cutorch} with \texttt{cudnnv4} library \\
\texttt{im2col+clBLAS} & \texttt{cltorch} (OpenCL Torch, this paper) \\
\bottomrule
\end{tabular}
\end{center}
\caption{Torch GPU Backends}
\label{tabletorchbackends}
\end{table}

\subsection{Convolutional Algorithms}\label{convolutional-algorithms}

In order to go further with hardware-agnostic deep learning frameworks,
it could be interesting to analyse how the CUDA learning frameworks
obtain excellent performance.

\needspace{48pt} There are four algorithms in common use:

\begin{itemize}
\tightlist
\item
  \texttt{im2col}
\item
  \texttt{implicitgemm}
\item
  \texttt{fft}
\item
  \texttt{winograd}
\end{itemize}

In addition, cuda-convnet\cite{cudaconvnet} provides a very fast direct
convolutional implementation for very specific convolutional geometries.

Let's look at each of these algorithms in turn, starting with
cuda-convnet

\subsubsection{Cuda-convnet}\label{cuda-convnet}

Cuda-convnet provides a very fast convolutional implementation for very
specific convolutional geometries. Unfortunately it generalizes poorly
to other geometries, or hardware. For example \cite{caffe} found that
using \texttt{im2col} gave a speedup of 1.3 times on K20, compared to
cuda-convnet, which is optimized for GTX580.

\subsubsection{im2col}\label{im2col}

\texttt{im2col} \cite{caffe,im2colchellapilla} is a very general method
for implementing convolutions, off-loading the geometry-specific
specializations to the underlying GPU BLAS implementation.
\texttt{im2col} converts the convolution of a single 3d image tensor
with the 4d weights tensor into a matrix multiplication of two
2-dimensional matrices. This uses more memory than direct convolution,
but it generalizes well to many convolutional geometries. It can take
advantage of existing optimizations in the existing SGEMM
implementations. The idea for using matrix multiplications for carrying
out convolutions is not new, but the observation that this produces
excellent performance on GPUs, that generalizes well to many geometries,
was made independently by \cite{caffe} and \cite{im2colchellapilla}.

For \texttt{im2col} in \texttt{cltorch}, the BLAS implementation is
handled by clBLAS\cite{clblas}. clBLAS is highly optimized for many
high-performance computing workloads, but till recently had not been
used for deep learning im2col workloads. The geometries produced by
im2col do not fit within the regime of geometries for which clBLAS has
been optimized on previously. Therefore, the SGEMM performance for these
geometries is relatively less good. Gu et al\cite{openclcaffe} found
that the overall convolutional speed of clBLAS for these workloads is
not competitive with \texttt{cublas}. clBLAS works well for large,
square matrices, with dimensions of around 1024 or 4096 along the side,
and dimensions a multiple of 8 or 16. Gu et al showed that using batched
\texttt{im2col}, over multiple images, puts clBLAS into a more favorable
regime, giving a performance boost of around 4-5 times, on AMD hardware.

Therefore it could be interesting to incorporate this technique
generally into other OpenCL backends.

In general, the performance of \texttt{im2col} will be strongly
dependent on the performance of the underlying \texttt{GEMM}
implementation.

\subsubsection{implicitgemm}\label{implicitgemm}

Similar to \texttt{im2col}, \texttt{implicitgemm}\cite{cudnn} lowers the
convolutions onto matrix multiplication. Whereas \texttt{im2col} fully
materializes the lowered matrix in off-chip memory,
\texttt{implicitgemm} materializes the lowered matrix lazily, in on-chip
memory. This reduces memory requirements, and can improve performance.
\texttt{implicitgemm} is implemented in NVIDIA's
\texttt{CUDNNv1}\cite{cudnn}.

An OpenCL implementation of \texttt{implicitgemm} is being developed by
Tschopp \cite{greentea}. This looks like a promising way forward to
improve OpenCL convolutional performance.

\subsubsection{FFT}\label{fft}

\texttt{fft} transforms the convolutional problem from the spatial
domain into the frequency domain. It is constant-time with respect to
the size of the convolutional kernel. It typically requires a large
amount of memory.\\
For models such as Alexnet, which have large kernels in the first layer,
\texttt{fft} gives noticeable speed benefits. Specifically, in the
second layer of AlexNet, \texttt{fft} shows a clear benefit for the
stride 1 5x5 kernels. However, recently the focus of convolutional
networks has turned to smaller kernels. VGG uses 3x3 kernels. GoogleNet
includes 1x1 kernels. On these models, the benefits of \texttt{fft} are
less clear, compared to other algorithms.

A competitive implementation of \texttt{fft} is
\texttt{fbfft}\cite{fbfft}. In Figure \ref{convnetbenchmarks}, it can be
seen that \texttt{fbfft} is fast on AlexNet, and on Overfeat, but offers
no advantage on VGG. No timings are available for GoogleNet, which
contains 1x1 convolutions, and is far from the favorable regime for
\texttt{fft}.

As for \texttt{im2col}, \texttt{fft} needs a fast \texttt{GEMM}
implementation. The performance of \texttt{fft} will be strongly
dependent on the speed of the underlying \texttt{GEMM}.

\subsubsection{Winograd}\label{winograd}

The Winograd FIR algorithms\cite{winogradarithmetic} are an algorithmic
optimization that reduces the number of mathematical operations required
for convolution. Winograd outlined the general principles in the 80s.
Lavin and Gray provided a specific implementation for GPU convolution in
\cite{lavin2015fast}, and implemented these algorithms in
\texttt{neon}\cite{neon}. \texttt{winograd} is an algorithmic
improvement, and therefore could be implemented also in OpenCL,
plausibly providing a speedup similar to the speedup which it provides
to \texttt{neon}. Note however that this is not the only optimization in
\texttt{neon}. We will look later at \texttt{neon}'s use of SASS,
including for the GEMM implementation, which provides additional
speedups.

As for \texttt{im2col} and \texttt{fft}, \texttt{winograd} transforms
the problem, but still relies on GEMM. And therefore its performance is
strongly related also to the efficiency of the underlying GEMM
implementation.

Having touched on \texttt{neon}, we should look at \texttt{neon}
specifically, as the current state of the art for convolutional
implementations, on NVIDIA CUDA hardware.

\subsubsection{Neon}\label{neon}

\texttt{neon}\cite{neon} uses at least the following optimizations to
obtain its very fast, near-optimal convolutional performance:

\begin{itemize}
\tightlist
\item
  \texttt{winograd} algorithm\cite{winogradarithmetic,lavin2015fast}
\item
  \texttt{SASS} implementation
\item
  \ldots{} including a \texttt{SASS} implementation of GEMM
\item
  fully fused kernels
\end{itemize}

As discussed, the \texttt{winograd} algorithm could be implemented in
OpenCL, and looks a promising possible way forward to improving OpenCL
convolutional performance. Note that as for \texttt{fft} and
\texttt{im2col}, \texttt{winograd} is also dependent on an underlying
GEMM implementation. Therefore, whilst writing \texttt{winograd} in
OpenCL could plausibly produce an efficient implementation, an efficient
implementation of GEMM would also be required.

The SASS implementations are not directly accessible from OpenCL, since
they are not just vendor-specific, but architecture specific. SASS
written for CUDA Maxwell devices is incompatible with SASS written for
Kepler devices, and visa versa. For example, the SASS implementations
make assumptions about the available registers. SPIR-V provides a
portable IL language, which could help obtain efficient optimizations.
However SPIR-V is at the level of PTX, rather than at the level of SASS.
PTX is higher-level than SASS. It is device independent IL, albeit
vendor-specific.

\begin{figure}[htbp]
\centering
\includegraphics[width=0.48000\textwidth]{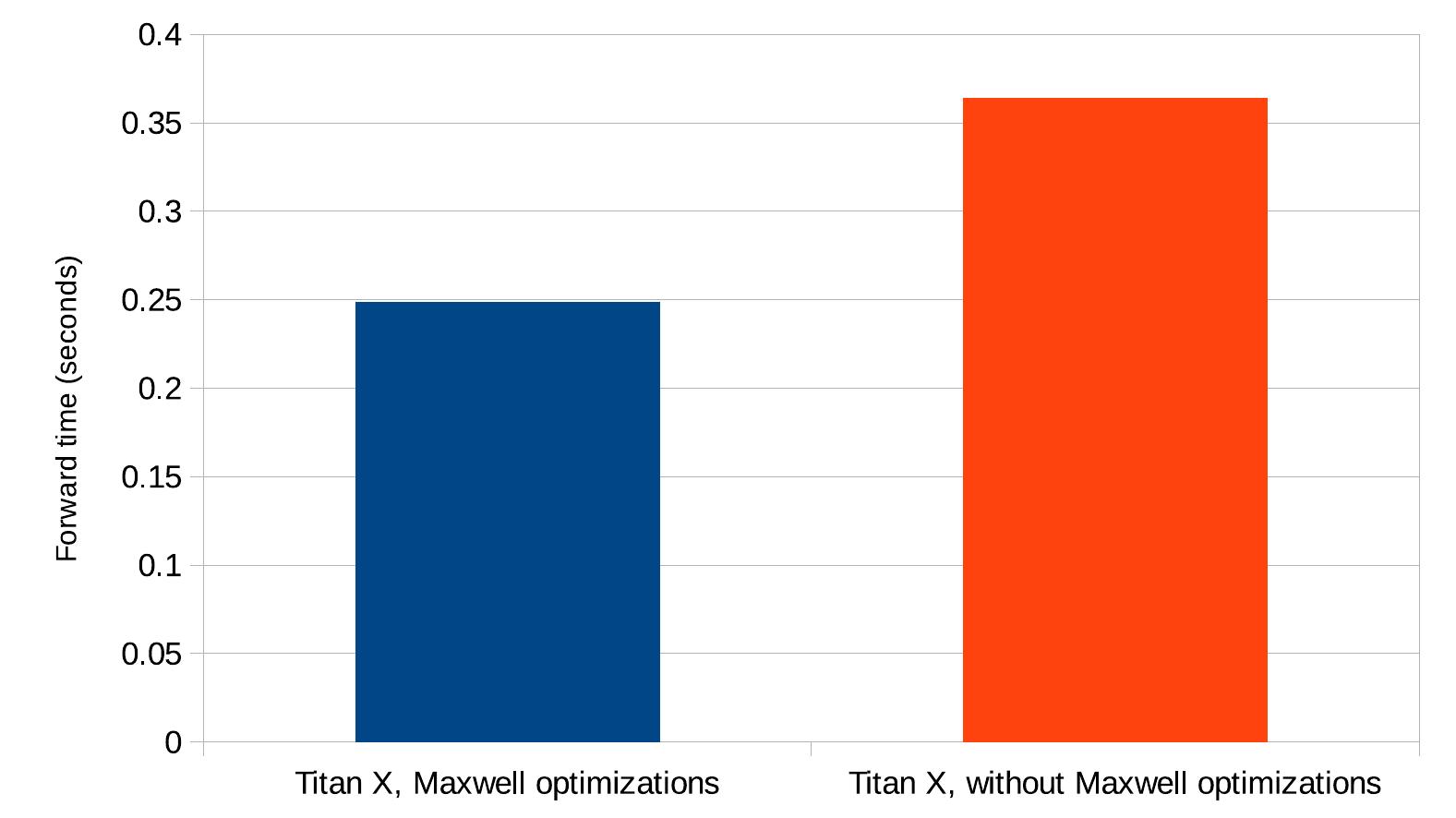}
\caption{Neon, effect of Maxwell
optimizations\label{neonmaxwellvskepler}}
\end{figure}

How much contribution does the SASS implementation, the custom GEMM, and
the fused kernels provide? We can estimate this by running a convolution
twice, on the same device, one with the Maxwell optimizations on, and
one with them turned off, by modifying the code to incorrectly detect
the device as Kepler. Figure \ref{neonmaxwellvskepler} shows the results
for this. This is for forward propagation of a batch of 128 224x224
images, using 3x3 convolutions, running on an NVIDIA Titan X. We can see
that the Maxwell optimizations reduce the batch time by about 33\%.

Therefore, an OpenCL implementation of \texttt{winograd} would be at
least around 33\% slower than a hardware-optimized low-level
implementation, such as \texttt{neon}. Taking into account the earlier
experiments of running ``Apply'' in OpenCL and in CUDA, on the same
device, which showed an additional performance drop of around 33\%, all
other things equal, we could expect an OpenCL implementation of
\texttt{winograd} to approach around 40\% of the execution speed of
\texttt{neon}.

However, note that \texttt{neon} depends also on the speed of the
underlying GEMM implementation, as do also \texttt{im2col} and
\texttt{fft}, so we should discuss GEMM briefly.

\subsection{GEMM}\label{gemm}

GEMM is at the heart of convolution. It is used by all the convolution
algorithms detailed above, with the exception of the direct convolution
algorithm in cuda-convnet. Thus, it is important that the GEMM
implementation should be the most efficient possible. Currently,
\texttt{cltorch} uses the clBLAS GEMM implementation. Gu et al showed
that clBLAS is an effective GEMM implementation that can be competitive
with the CUDA \texttt{cublas} implementation. Gu et al showed that it is
important to ensure that the matrix sizes fall close to the optimal
regime for clBLAS. Gu et al showed that by using batching, lower
multiple images into matrix multiplication in a single batch, the clBLAS
implementation was around 4-5 times faster, on AMD hardware.

A possible alternative to clBLAS is ViennaCL, which provides a highly
hardware-agnostic implementation of GEMM, working not just on OpenCL,
but also on CUDA and OpenMP.

\begin{figure}[htbp]
\centering
\includegraphics[width=0.41000\textwidth]{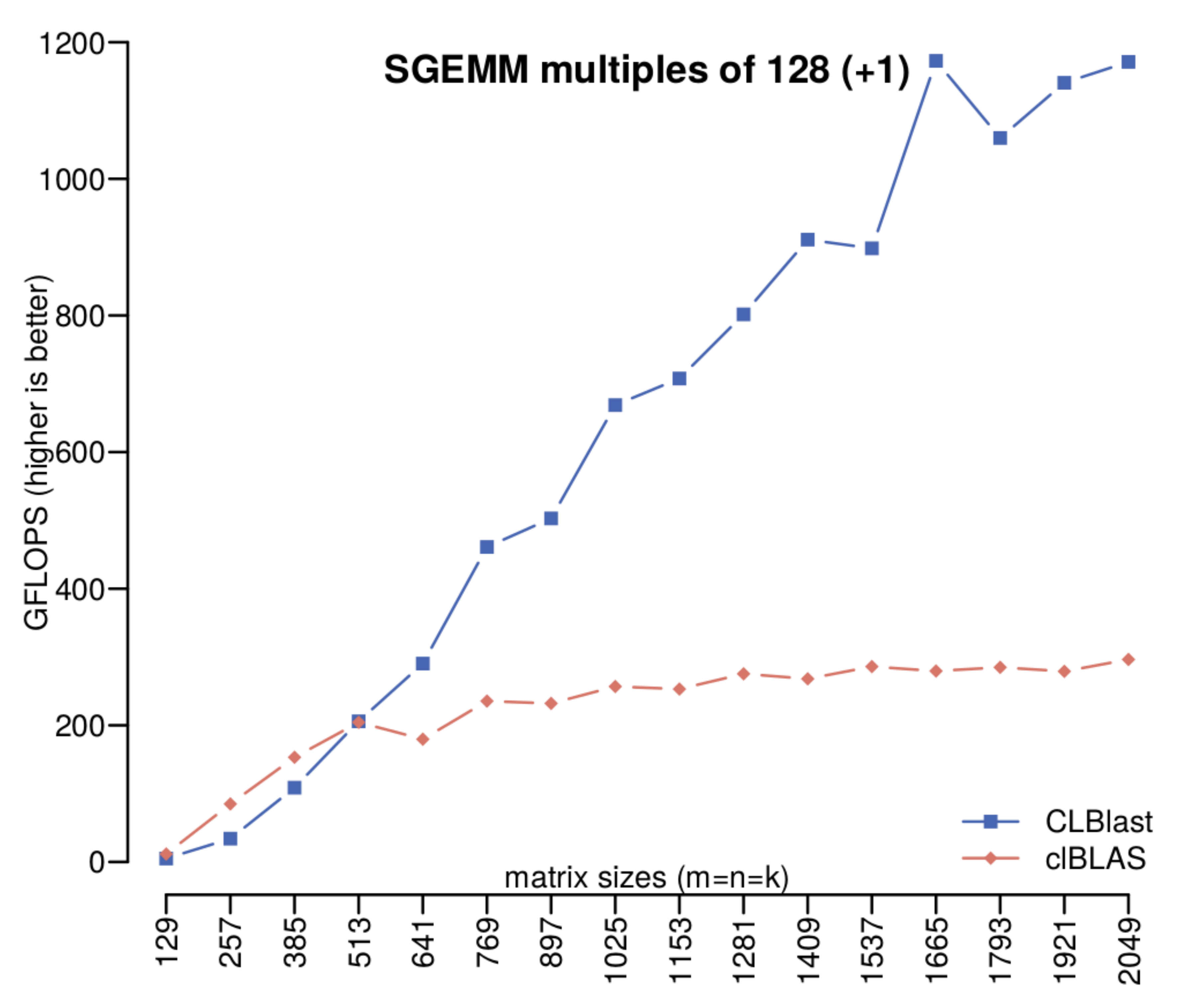}
\caption{Comparison of CLBlast with clBLAS, on Radeon
HD7950\label{graph:clblast}}
\end{figure}

In terms of performance, Nugteren is developing CLBlast\cite{clblast},
based on CLTune auto-tuner\cite{cltune}. CLBlast shows a performance
benefit relative to clBLAS for matrices whose sides are not a multiple
of \(128\). Specifically matrices with a side of \(128m+1\), for integer
\(m\), show a clear benefit, Figure \ref{graph:clblast}, \cite{clblast}.
Tschopp's \texttt{implicitgemm} implementation is based loosely on the
CLBlast GEMM implementation, but using fully fused kernels, rather than
factorizing the matrix lowering operation and the GEMM into separate
kernels.

It could be interesting to benchmark OpenCL \texttt{GEMM}
implementations, under workloads associated with the convolutional
algorithms discussed above. Lokhmotov's GEMMbench \cite{gemmbench} could
potentially be useful for this.

Going back to \texttt{neon}, assuming that one could obtain a
\texttt{GEMM} implementation competitive with \texttt{cublas} for
OpenCL, then it looks like it could be possible to write an OpenCL
implementation of \texttt{winograd}, that could approach around 20-30\%
of the speed of \texttt{neon} (ie 70-80\% slower than \texttt{neon}).
This would be a significant improvement on the current \texttt{im2col}
implementations. However, can we do better?

\subsection{Pluggable Implementations}\label{pluggable-implementations}

It seems challenging to approach the performance of close-to-the-metal
\texttt{SASS} from within OpenCL or SPIR-V, because there is a limit to
how far one can improve the performance in languages designed to be
portable. To go further, two possible approaches could be:

\begin{itemize}
\tightlist
\item
  implementation of highly optimizing compilers, for OpenCL, or
\item
  create pluggable hardware-specific convolutional implementations
\end{itemize}

The former approach, of creating highly optimizing compilers, is an
active area of research, and there are no easy answers. AEcute metadata
\cite{aecutemetadata}, and the more recent PENCIL\cite{pencil} are two
approaches to facilitate generation of hardware-optimized code. PENCIL
allows expression of algorithms in a higher-level language, which can be
used to generate device-optimized OpenCL.

Rather than attempting to make the convolutional implementations
portable, an alternative approach could be to make them pluggable,
loadable at runtime, and strongly hardware-specific. Thus they could be
written in low-level assembler, and make full use of hardware-specific
optimizations, such as knowledge of the exact memory dimensions,
register layout, and other hardware characteristics The neural network
library frameworks could themselves be written in a portable language,
such as OpenCL.

\begin{figure}[htbp]
\centering
\includegraphics[width=0.50000\textwidth]{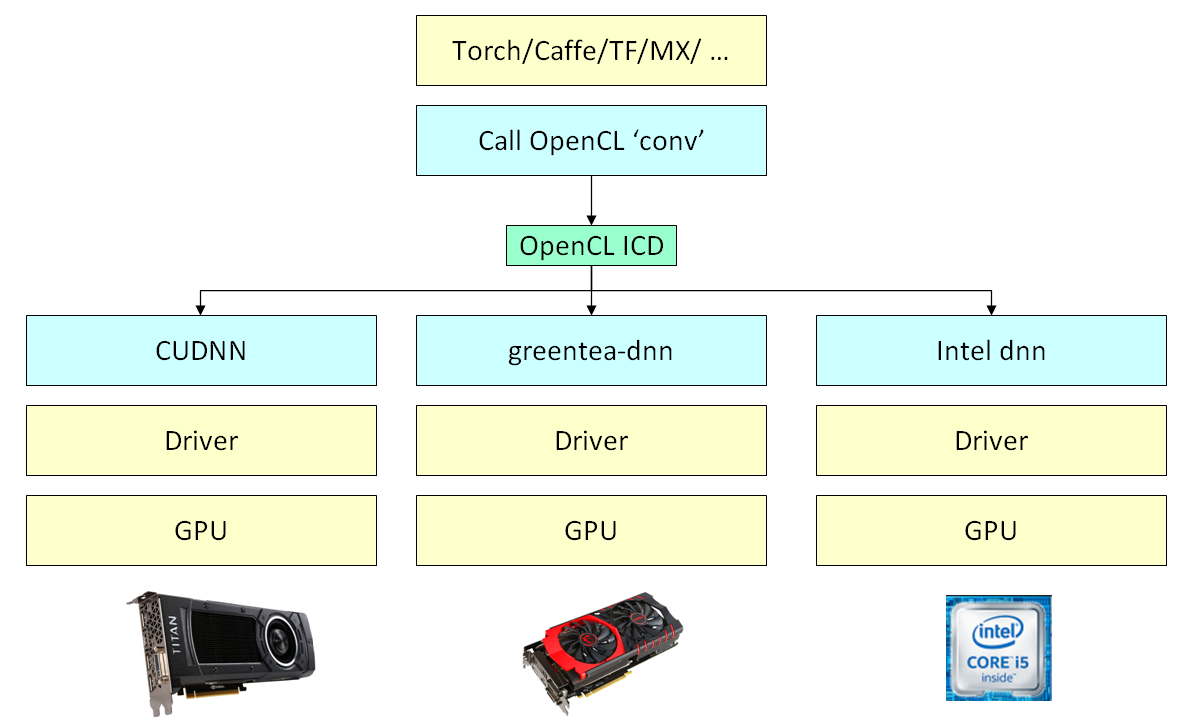}
\caption{Proposal for runtime convolution linking\label{convprop}}
\end{figure}

Concretely, this could work in a similar way to the \texttt{ICD}
abstraction layer in OpenCL. OpenCL's \texttt{ICD} binds with the
vendor-provided OpenCL implementation at runtime. In the case of
convolution, the machine learning framework could simply call a function
\texttt{conv}, within a Khronos-provided API. The Khronos API would
route the call to a convolutional implementation appropriate to the
current targeted hardware device. This might look something like Figure
\ref{convprop}.

\begin{figure}[htbp]
\centering
\includegraphics[width=0.50000\textwidth]{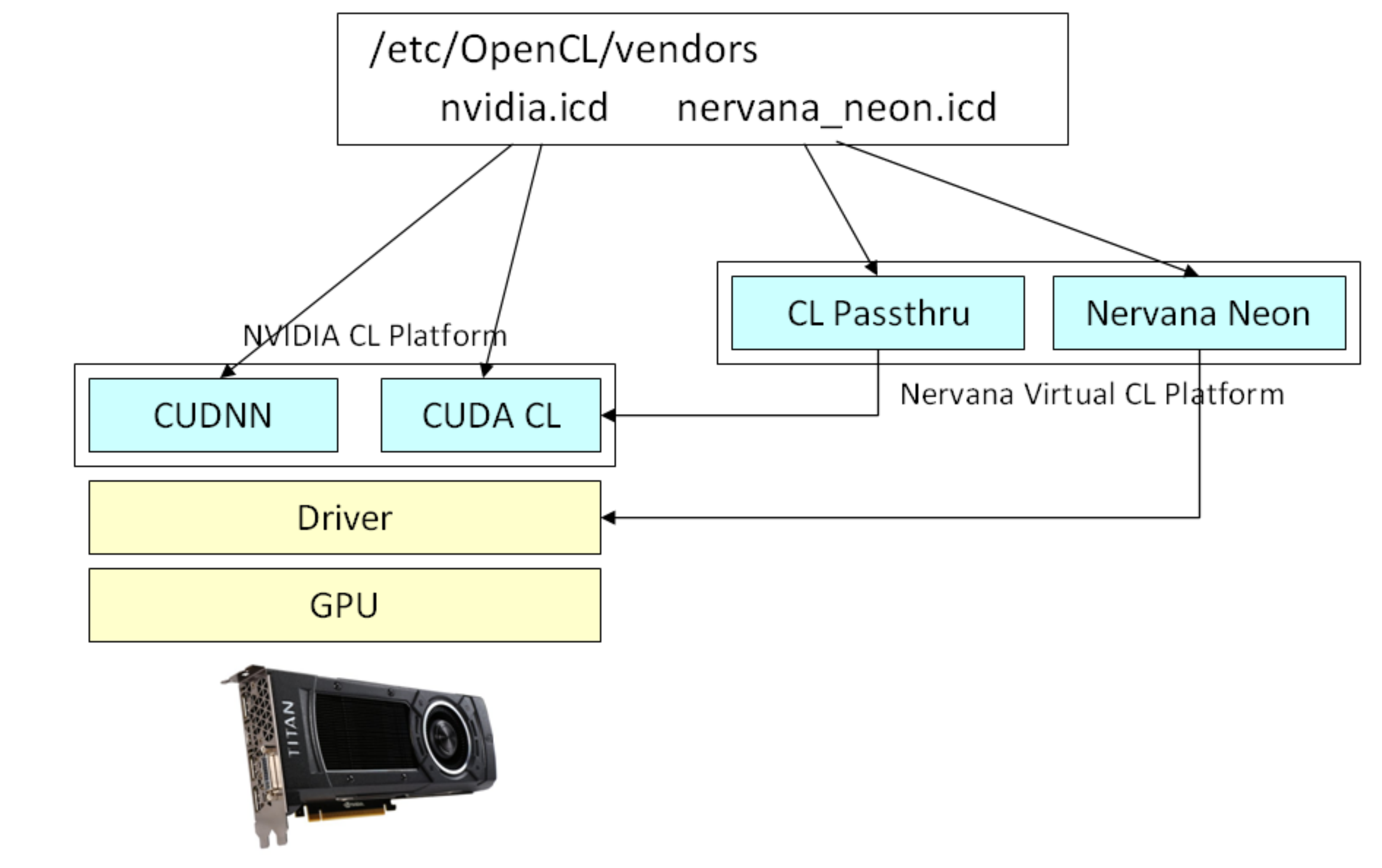}
\caption{Proposal for virtual OpenCL platform\label{convprop2}}
\end{figure}

Note that this architecture says nothing about who will write the highly
optimized hardware-specific convolutional implementation. For example we
could imagine a virtual OpenCL platform, that wraps the \texttt{neon}
implementation, for CUDA hardware, see Figure \ref{convprop2}.

\subsection{HCC}\label{hcc}

Discussion of non-CUDA deep learning implementations would not be
complete without touching on AMD's HCC. HCC is an alternative open
standard to OpenCL. It is not backed by Khronos, but it is an open
standard, backed by one of the major competitors to NVIDIA. It is
currently implemented only on AMD hardware, but there is no reason why
other hardware vendors couldnt support it in the future. Since only one
vendor owns the specifications currently, it can potentially evolve
rapidly.

Note that HCC will not in itself directly solve the challenge of writing
a fast convolutional layer. The key advantages of HCC could be agility,
and compatibility with CUDA language. The AMD implementation of HCC
might become faster than the AMD implementation of OpenCL. However, it
seems unlikely that convolutional implementations written in pure HCC
would approach the performance of Gray's SASS implementations in
\texttt{neon}. Therefore, the use of pluggable architectures, such as
that outlined above, might be key to obtaining the fastest
cross-platform convolutional performance.

HCC solves many of the challenges faced during the development of
\texttt{cltorch}, such as the use of c++ templates in the \texttt{torch}
CUDA code. It could therefore significantly facilitate the development
of hardware- agnostic deep learning frameworks in the future. The extent
to which HCC can contribute to hardware-agnostic deep learning framework
will depend on the extent to which multiple hardware vendors adopt the
standard.

\section{Possible future evolutions}\label{possible-future-evolutions}

In previous sections, possible future evolutions for portable neural
network compute have been described in detail. Summarizing here, several
possible future evolutions could improve the performance of portable
deep learning libraries:

\begin{itemize}
\tightlist
\item
  OpenCL implementation of \texttt{implicitgemm}, for example that
  currently under development by Tschopp
\item
  OpenCL implementation of Winograd algorithms
\item
  Improvements to the OpenCL GEMM implementations, for example that
  currently under development by Nugteren
\item
  hardware-specific convolutional implementations, loaded at runtime via
  a pluggable architecture
\end{itemize}

Finally, HCC might facilitate the future development of
hardware-agnostic learning frameworks, by facilitating the porting
process, and reducing the disparity between CUDA and non-CUDA
code-bases.

\section{Conclusion}\label{conclusion}

\texttt{cltorch} provides a hardware-agnostic backend for the Torch deep
learning library. It is written using the portable language `OpenCL'.
\texttt{cltorch} provides sufficient functionality to train recent
models such as: AlexNet, GoogleNet, Overfeat, and VGG. Challenges faced
during the creation of the \texttt{cltorch} backend were presented.
Current OpenCL neural network framework implementations face challenges
obtaining training speeds competitive with CUDA implementations.
Possible approaches to improve the speed of the convolutional
implementation have been presented. Finally, HCC could solve many of the
challenges faced during the development and reduce the disparity between
CUDA and non-CUDA codebases, but the extent to which it will become a
hardware-agnostic standard, implemented by multiple vendors is as yet
uncertain.

\bibliographystyle{abbrv}
\bibliography{cltorch2}

\end{document}